\documentclass[letterpaper, 10 pt, conference]{cls/ieeeconf}  

\IEEEoverridecommandlockouts                              

\makeatletter
\newcommand{\linebreakand}{%
  \end{@IEEEauthorhalign}\hfill\mbox{}\par\mbox{}\hfill\begin{@IEEEauthorhalign}}
\makeatother

\usepackage[table,dvipsnames]{xcolor}      
\usepackage[noadjust]{cite}

\usepackage{amsmath,amssymb,amsfonts,amsthm,dsfont,mathtools}
\usepackage{graphicx,tabularx,adjustbox}
\graphicspath{{./fig/}}
\usepackage[font=small]{caption}
\usepackage[font=footnotesize]{subcaption}
\usepackage{balance}
\usepackage[breaklinks=true, colorlinks, bookmarks=true]{hyperref}

\newcommand{\pa}{\partial}

\newtheorem{proposition}{Proposition}
\newtheorem{lemma}{Lemma}

\theoremstyle{definition}
\newtheorem{definition}{Definition}
\newtheorem*{problem*}{Problem}

\newtheorem*{remark*}{Remark}

\newcommand{\bfa}{\mathbf{a}}
\newcommand{\bfb}{\mathbf{b}}
\newcommand{\bfc}{\mathbf{c}}
\newcommand{\bfd}{\mathbf{d}}

\newcommand{\bfg}{\mathbf{g}}

\newcommand{\bfn}{\mathbf{n}}

\newcommand{\bfp}{\mathbf{p}}
\newcommand{\bfq}{\mathbf{q}}

\newcommand{\bfu}{\mathbf{u}}
\newcommand{\bfv}{\mathbf{v}}

\newcommand{\bfx}{\mathbf{x}}
\newcommand{\bfy}{\mathbf{y}}
\newcommand{\bfz}{\mathbf{z}}

\newcommand{\bfrho}{\boldsymbol{\rho}}

\newcommand{\bfC}{\mathbf{C}}

\newcommand{\bbR}{\mathbb{R}}

\newcommand{\calD}{\mathcal{D}}

\newcommand{\calF}{\mathcal{F}}

\newcommand{\calK}{\mathcal{K}}
\newcommand{\calL}{\mathcal{L}}

\newcommand{\calO}{\mathcal{O}}
\newcommand{\calP}{\mathcal{P}}

\newcommand{\calS}{\mathcal{S}}

\newcommand{\calU}{\mathcal{U}}

\newcommand{\calX}{\mathcal{X}}

\newcommand{\calZ}{\mathcal{Z}}

\title{\LARGE \bf
Control Strategies for Pursuit-Evasion Under Occlusion Using Visibility and Safety Barrier Functions
}

\author{Minnan Zhou$^{1,*}$ \and Mustafa Shaikh$^{1,*}$ \and Vatsalya Chaubey$^{1,*}$ \and 
Patrick Haggerty$^2$ \linebreakand Shumon Koga$^3$ \and Dimitra Panagou$^4$ \and Nikolay Atanasov$^1$%
\thanks{This work was supported by ARL DCIST grant W911NF-17-2-0181.}%
\thanks{$^*$These authors contributed equally.}%
\thanks{$^1$The authors are with the Department of Electrical and Computer Engineering, University of California San Diego, La Jolla, CA 92093, USA (e-mails: {\tt\footnotesize \{m9zhou,mushaikh,vchaubey,natanasov\}@ucsd.edu}).}%
\thanks{$^2$Patrick Haggerty is with General Dynamics Mission Systems, Bloomington, MN 55431, USA (e-mail: {\tt\footnotesize patrick.haggerty@gd-ms.com}).}%
\thanks{$^3$Shumon Koga is with Honda R\&D Co. Ltd., Tokyo, Japan (e-mail: {\tt\footnotesize shumon\_koga@jp.honda}).}%
\thanks{$^4$Dimitra Panagou is with the Department of Robotics and the Department of Aerospace Engineering, University of Michigan, Ann Arbor, MI 48109, USA (e-mail: {\tt\footnotesize dpanagou@umich.edu}).}%
\thanks{\url{https://existentialrobotics.org/VisibilityControl}}
}

\begin{document}

\maketitle
\thispagestyle{empty}
\pagestyle{empty}

\begin{abstract}
This paper develops a control strategy for pursuit-evasion problems in environments with occlusions. We address the challenge of a mobile pursuer keeping a mobile evader within its field of view (FoV) despite line-of-sight obstructions. The signed distance function (SDF) of the FoV is used to formulate visibility as a control barrier function (CBF) constraint on the pursuer's control inputs. Similarly, obstacle avoidance is formulated as a CBF constraint based on the SDF of the obstacle set. While the visibility and safety CBFs are Lipschitz continuous, they are not differentiable everywhere, necessitating the use of generalized gradients. To achieve non-myopic pursuit, we generate reference control trajectories leading to evader visibility using a sampling-based kinodynamic planner. The pursuer then tracks this reference via convex optimization under the CBF constraints. We validate our approach in CARLA simulations and real-world robot experiments, demonstrating successful visibility maintenance using only onboard sensing, even under severe occlusions and dynamic evader movements.
\end{abstract}


\section{INTRODUCTION}
\label{sec:introduction}

Pursuit-evasion problems \cite{chung2011search} are studied in computational geometry, control theory, and robotics, motivated by applications in search and rescue \cite{kumar2004robot}, security and surveillance \cite{grocholsky2006cooperative}, and environmental monitoring \cite{julian2019distributed}. Introducing visibility constraints leads to the art gallery problem \cite{art_gallery}, which has elegant solutions with static pursuers (guards) in 2D polygonal environments but becomes challenging with mobile pursuers in 3D environments. Prior works tackled pursuit-evasion with visibility constraints by graph-theoretic \cite{gerkey2006visibility} or game-theoretic approaches \cite{bhattacharya2009game,bhattacharya2010existence, ZouVisibilityTrackingGame}. This work proposes a novel control design for pursuit-evasion with field-of-view (FoV) constraints, using control barrier function (CBF) techniques.

Originating from the pioneering works of Wieland and Allg{\"o}wer \cite{wieland2007cbf} and Ames et al. \cite{ames2016control,ames2019control}, CBFs have become a key tool for safety-critical applications such as adaptive cruise control \cite{AmesACC}, robot manipulation \cite{cortez2019controlbarrierfunctionsmechanical}, robot locomotion \cite{GrandiaLeggedRobots}, and robot flight \cite{wang2017safecertificatebasedmaneuversteams}. CBFs are defined so that the positivity of a barrier function implies forward invariance of a corresponding superlevel set for the trajectories of a dynamical system. A key observation is that, for control-affine dynamical systems, CBF constraints are \emph{linear} in the control input. This allows control synthesis via quadratic programming (QP) subject to linear CBF constraints \cite{ames2019control}.

\begin{figure}[t]
    \centering
    \includegraphics[width=0.9\linewidth]{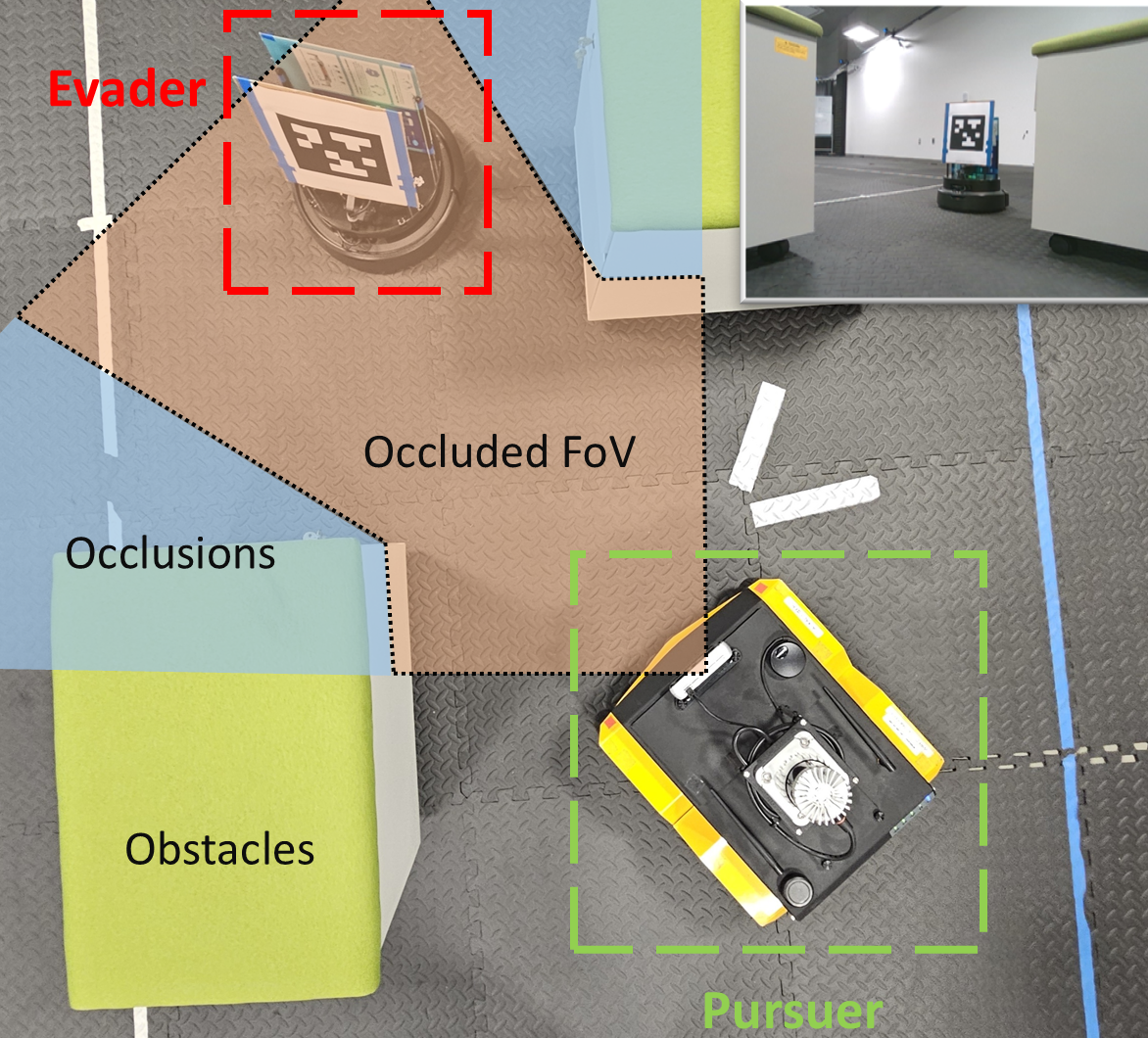}
    \caption{A mobile \emph{pursuer} aims to keep a mobile \emph{evader} within its field of view despite occlusions and without colliding with obstacles. The inset shows the evader in the pursuer's camera view.}
    \label{fig:system}
\end{figure}

A challenge for CBF methods is to determine a function whose positivity is equivalent to the desired safe behavior. Previous works have demonstrated ways to validate candidate CBFs for polynomial control systems \cite{clark2021verificationsynthesiscontrolbarrier} subject to multiple state and input constraints \cite{IsalyFeasibility}, or even to learn CBFs from expert demonstrations  \cite{robey2020learning}. A key observation in this paper is that visibility under occlusion can also be formulated as a CBF condition using the signed distance function (SDF) of the sensor FoV. An SDF provides the (signed) distance from any point in space to a given set's boundary. SDFs have been used to model shapes and surfaces in vision and robotics tasks such as object shape reconstruction \cite{park2019deepsdf} and robot mapping \cite{oleynikova2017voxblox}. In this paper, we introduce a novel \emph{visibility CBF constraint} that enforces negativity of the SDF from an evader to the occluded FoV of the pursuer. 

While early works on CBFs primarily focused on continuous systems subject to differentiable CBF constraints, recent advances \cite{Glotfelter_non_smooth_barrier, Ghanbarpour_non_smooth_differential_inclusions, IsalyFeasibility} have studied the regularity of CBF-based controllers and have expanded their applicability to non-smooth dynamics and non-smooth constraints. These methods have been used in \cite{thirugnanam2023nonsmoothcontrolbarrierfunctions} to enable safe multi-robot navigation with polytope-shaped robots. In our case, the visibility CBF is Lipschitz continuous but not differentiable everywhere. It is also time-varying due to the pursuer's and evader's motion. Hence, it is necessary to use a non-smooth time-varying formulation of the visibility CBF constraint involving its generalized gradient.

Maintaining visibility is a requirement in various target tracking problems.  A visibility-aware planner proposed in \cite{Wang_visibility_aware_traj_planning} generates sensor trajectories with maximal visibility to a moving target. A vision-based controller for a quadrotor landing on a ground vehicle, while preventing camera occlusions, was developed in \cite{hoang2017vision}. Leader-follower formations for maintaining visibility and safety were developed using dipolar vector fields in a known environment in \cite{panagou2012maintaining}. Similarly, in the presence of FoV constraints, dipolar vector fields have been used in \cite{maniatopoulos2013model} to formulate a model predictive control scheme for differential-drive robot navigation, maintaining visibility of a static landmark. Closely related to our work, Gao et al. \cite{gao2023probabilistic} formulate a probabilistic notion of visibility under occlusion, which is maintained using an extended Kalman filter \cite{kalmanbucy1961}. The authors develop a real-time non-myopic trajectory planner for visibility-aware and safe target tracking under uncertainty. This work is complementary to ours in that it focuses on high-level non-myopic visibility planning using nonlinear optimization and linear SDF approximations, while we focus on low-level myopic visibility control to track a planned reference using convex programming. The contributions of our work are summarized,
\begin{itemize}
    \item We prove Lipschitz continuity of the signed distance to an occluded FoV to justify its use as a non-smooth time-varying CBF for visibility maintenance.
    
    \item We achieve non-myopic visibility maintenance by coupling a kinodynamic planner that generates reference controls leading to evader visibility, with a controller that tracks the planned reference using convex optimization subject to visibility and safety CBF constraints.

    \item We demonstrate successful visibility maintenance in real robot experiments using only onboard sensing, even under severe occlusions and dynamic evader motion.
\end{itemize}

\section{PROBLEM STATEMENT}
\label{sec:background}

Consider a mobile robot modeled as a system with state $\bfx \in \bbR^n$, control input $\bfu \in \calU \subset \bbR^m$, and dynamics model:
\begin{align} \label{eq:pursuer} 
    \dot{\bfx}(t) = f(\bfx(t)) + G(\bfx(t)) \bfu(t), 
\end{align}
where $f: \bbR^n \to \bbR^n$ and $G: \bbR^n \to \bbR^{n \times m}$ are continuous functions. We refer to the robot as a \emph{pursuer} because we are interested in controlling its motion to maintain visibility of an \emph{evader}. The evader is modeled as a system with configuration $\bfy(t) \in \bbR^p$ and velocity $\dot{\bfy}(t) \in \bbR^p$. The evader's velocity is assumed to be bounded, i.e., $\|\dot{\bfy}(t)\| \leq k$ for some $k < \infty$.

The pursuer is equipped with a sensor with \emph{unoccluded} field of view (FoV), i.e., in the absence of obstacles, modeled by a closed set $\calF_0 \subset \bbR^p$, determined by the onboard sensor ranges and parameters. The unoccluded FoV at a particular pursuer state $\bfx$, i.e., transformed to the global coordinate frame, is represented as a linear transformation, $A(\bfx)\calF_0 + b(\bfx)$, where $A: \bbR^n \rightarrow \bbR^{p \times p}$ and $b: \bbR^n \rightarrow \bbR^p$ are Lipschitz continuous functions and the multiplication is defined for every element of $\calF_0$. For example, if $\bfx$ contains the position $\bfp$ and orientation $R$ of the pursuer, the transformed FoV would be $R\calF_0 + \bfp$. We denote the \emph{occluded} FoV, i.e., determined by obstacles in the environment depending on the pursuer's state $\bfx$, by $\calF(\bfx) \subseteq A(\bfx)\calF_0 + b(\bfx)$. The evader is \emph{visible} to the pursuer at time $t$ if $\bfy(t) \in \calF(\bfx(t))$.

Our objective is to design a control policy for the pursuer to maintain visibility of the evader. In addition, the pursuer's state $\bfx(t)$ should remain outside of an obstacle set $\calO \subset \bbR^n$ at all time. In summary, we consider the following problem.

\begin{problem*}
Let $\bfx(0) \notin \calO$ be an initial condition for the pursuer system in \eqref{eq:pursuer}. Given the evader state $\bfy(t)$, $\dot \bfy(t)$ for all $t \geq 0$, design a control policy for the pursuer that ensures $\bfx(t) \notin \calO$ for all $t \geq 0$ and maximizes the duration of evader visibility $\bfy(t) \in \calF(\bfx(t))$. 
\end{problem*}

We assume that the evader state $\bfy(t)$, $\dot{\bfy}(t)$ is known at time $t$ and focus on pursuer motion planning and control instead of on evader motion estimation. In practice, the evader state can be estimated using onboard sensing, e.g., via visual tracking \cite{Bewley_sort} for a camera-equipped pursuer.

\section{TECHNICAL APPROACH}
\label{sec:technical_approach}

Our key idea is to encode the requirements for the pursuer to maintain visibility and safety as CBF constraints, and synthesize pursuer control inputs subject to these constraints. We begin with a review of the concept of CBF.

\subsection{Control Barrier Functions}

In safety-critical applications, it is desirable to control a dynamical system to ensure its state remains in a safe set:
\begin{align} \label{prelim:eq:safeset} 
    \calS(t) &= \{ \bfx \in \bbR^n | \; h(t,\bfx) \geq 0\}, 
\end{align}
defined as the superlevel set of a continuously differentiable function $h: \bbR_+ \times \bbR^n \to \bbR$. 

\begin{definition}[{\cite{garg2021robust}}]
\label{def:cbf}
A function $h: \bbR_+ \times \bbR^n \to \bbR$ is a \emph{time-varying control barrier function} (CBF) on set $\calD$, if $\calS(t)$ in \eqref{prelim:eq:safeset} satisfies $\calS(t) \subset \calD \subseteq \bbR^n$ and there exists an extended class $\calK$ function $\alpha: \bbR \to \bbR$ such that 
\begin{equation*}
    \sup_{\bfu \in \calU} \left\{ \calL_{f} h(t,\bfx) + \calL_{G} h(t,\bfx) \bfu + \frac{\partial h}{\partial t}(t,\bfx) \right\}  \geq - \alpha (h(t,\bfx)),  
\end{equation*}
for all $\bfx \in \calD$ and for all $t \geq 0$. The terms $\calL_{f} h(t,\bfx) = \nabla_\bfx h(t,\bfx)^\top f(\bfx)$ and $\calL_{G} h(t,\bfx) = \nabla_\bfx h(t,\bfx)^\top G(\bfx)$ are the Lie derivatives of $h(t,\bfx)$ along the vector fields $f$, $G$ in \eqref{eq:pursuer}.
\end{definition}

When $\calS(t)$ is defined by a CBF, it can be guaranteed that the system state $\bfx(t)$ remains within $\calS(t)$ (invariance) using a CBF constraint on the system control inputs.

\begin{lemma}[{\cite{garg2021robust}}]
Let $h : \bbR_+ \times \bbR^n \to \bbR$ be a time-varying CBF defining a safe set \eqref{prelim:eq:safeset}. Then, any Lipschitz continuous control policy $\bfu = \pi(t,\bfx)$ for the system in \eqref{eq:pursuer} satisfying 
\begin{equation}\label{eq:CBF-condition}
\calL_{f} h(t,\bfx) + \calL_{G} h(t,\bfx) \pi(t,\bfx) + \frac{\partial h}{\partial t}(t,\bfx) \geq \!- \alpha (h(t,\bfx)),\!
\end{equation}
for all $\bfx \in \calS(t)$ and for all $t \geq 0$, renders $\calS(t)$ invariant.
\end{lemma} 

For a locally Lipschitz reference controller $r(t,\bfx)$, the quadratic program below with CBF safety constraint provides the minimum perturbation to $r(t,\bfx)$ to guarantee safety \cite{ames2019control}:
\begin{equation}\label{controller: CBF-QP}
\begin{aligned} 
\pi(t,\bfx) = \arg\min_{\bfu \in \calU} &\, \|\bfu - r(t,\bfx)\|^2\\
    \mathrm{s.t.} \;\; &\, \calL_{f} h + \calL_{G}h \bfu + \frac{\partial h}{\partial t} \geq - \alpha(h).
\end{aligned}
\end{equation}

Barrier functions in real-world applications are not necessarily smooth. As described in \cite{Glotfelter_non_smooth_barrier}, generalized gradients and regularity conditions can be used to deal with non-differentiable points of a locally Lipschitz CBF.

\begin{definition}[{\cite[Thm.~2.5.1]{clarke1990optimization}}]
\label{def:generalized_gradient}
Let $h$ be Lipschitz near $\bfz = (t,\bfx) \in \bbR^{n+1}$, and suppose $\calZ$ is any set of Lebesgue measure zero in $\bbR^{n+1}$. The generalized gradient of $h$ is
\begin{equation}
\partial h(\bfz)=\operatorname{co}\left\{\lim _{i \rightarrow \infty} \nabla h\left(\bfz_i\right) \mid \bfz_i \rightarrow \bfz,\bfz_i \notin \calZ \cup \Omega_h\right\},
\end{equation}
where $\Omega_h$ is the zero-measure set where $h$ is non-differentiable and $\operatorname{co}$ is the convex hull.
\end{definition}

\begin{definition}[{\cite[Def.~2.3.4]{clarke1990optimization}}]
A function $h$ is regular at $\bfz \in \bbR^{n+1}$ provided that for all $\bfd \in \mathbb{R}^{n+1}$, the one-sided directional derivative $h^{\prime}(\bfz ; \bfd)=\lim_{e \downarrow 0} e^{-1}(h(\bfz + e\bfd)-h(\bfz))$ exists and satisfies $h^{\prime}(\bfz ; \bfd) = \limsup _{\substack{\bfz' \rightarrow \bfz \\ e \downarrow 0}} \frac{h(\bfz'+ e\bfd)-h(\bfz')}{e}$.
\end{definition}

Following the result in \cite[Thm.~2]{Glotfelter_non_smooth_barrier}, we can extend the notion of time-varying CBF in Def.~\ref{def:cbf} to functions $h(t,\bfx)$ that are not differentiable everywhere.

\begin{definition}\label{nbf}
A locally Lipschitz, regular function $h(t,\bfx)$ is a \emph{non-smooth time-varying CBF} on a set $\calD$ if $\calS(t)$ in \eqref{prelim:eq:safeset} satisfies $\calS(t) \subset \calD \subseteq \bbR^n$, and there exists a locally Lipschitz extended class-$\mathcal{K}$ function $\alpha: \mathbb{R} \rightarrow \mathbb{R}$ such that:
\begin{equation*}
\sup_{\bfu \in \calU} \inf_{(w,\bfv) \in \partial h(t,x)} \{ \bfv^\top(f(\bfx)+G(\bfx)\bfu) + w\} \geq - \alpha(h(t,\bfx))
\end{equation*}
for all $\bfx \in \calS(t)$ and for all $t \geq 0$.
\end{definition}

\subsection{Visibility and Safety as CBF Constraints}

The condition that the evader is visible to the pursuer can be formulated as a non-smooth time-varying CBF. We model the safe set as the pursuer's FoV $\calF(\bfx)$, and the CBF as the signed distance from the evader $\bfy(t)$ to the FoV boundary.

\begin{definition}
The \emph{signed distance function} $d : \bbR^p \times 2^{\bbR^p} \to \bbR$ computes the signed distance from a point $\bfq \in \bbR^p$ to the boundary $\pa \calF$ of a set $\calF \subset \bbR^p$ as:
\begin{align}
     d(\bfq, \calF)  = 
     \begin{cases}
     - \min_{\bfq^* \in \pa \calF}  \| \bfq - \bfq^*\|, \quad \textrm{if} \quad \bfq \in {\mathcal F}, \\
     \phantom{-} \min_{\bfq^* \in \pa \calF} \|\bfq - \bfq^*\|, \quad \textrm{if} \quad \bfq \notin \calF. 
     \end{cases}
\end{align}
\end{definition}

We define a \emph{visibility CBF} using the signed distance to the pursuer's FoV:
\begin{equation}\label{eq:visibility_barrier}
h(t,\bfx) = -d(\bfy(t), \calF(\bfx)).
\end{equation}
Note that $h(t,\bfx)$ in \eqref{eq:visibility_barrier} is not differentiable at all points because the FoV can be an arbitrary set under occlusion such that $\calF(\bfx) \subseteq A(\bfx) \calF_0 + b(\bfx)$. However, we show that $h(t,\bfx)$ is Lipschitz continuous and thus a valid non-smooth time-varying CBF.

\begin{proposition}\label{proposition:visibility_lipschitz}
With bounded evader velocity, $\|\dot{\bfy}(t)\| \!\leq\! k$, the function $h(t, \bfx) = -d(\bfy(t), \calF(\bfx))$ is Lipschitz continuous.
\end{proposition}

\begin{figure}[t]
    \centering
    \includegraphics[width=0.9\linewidth]{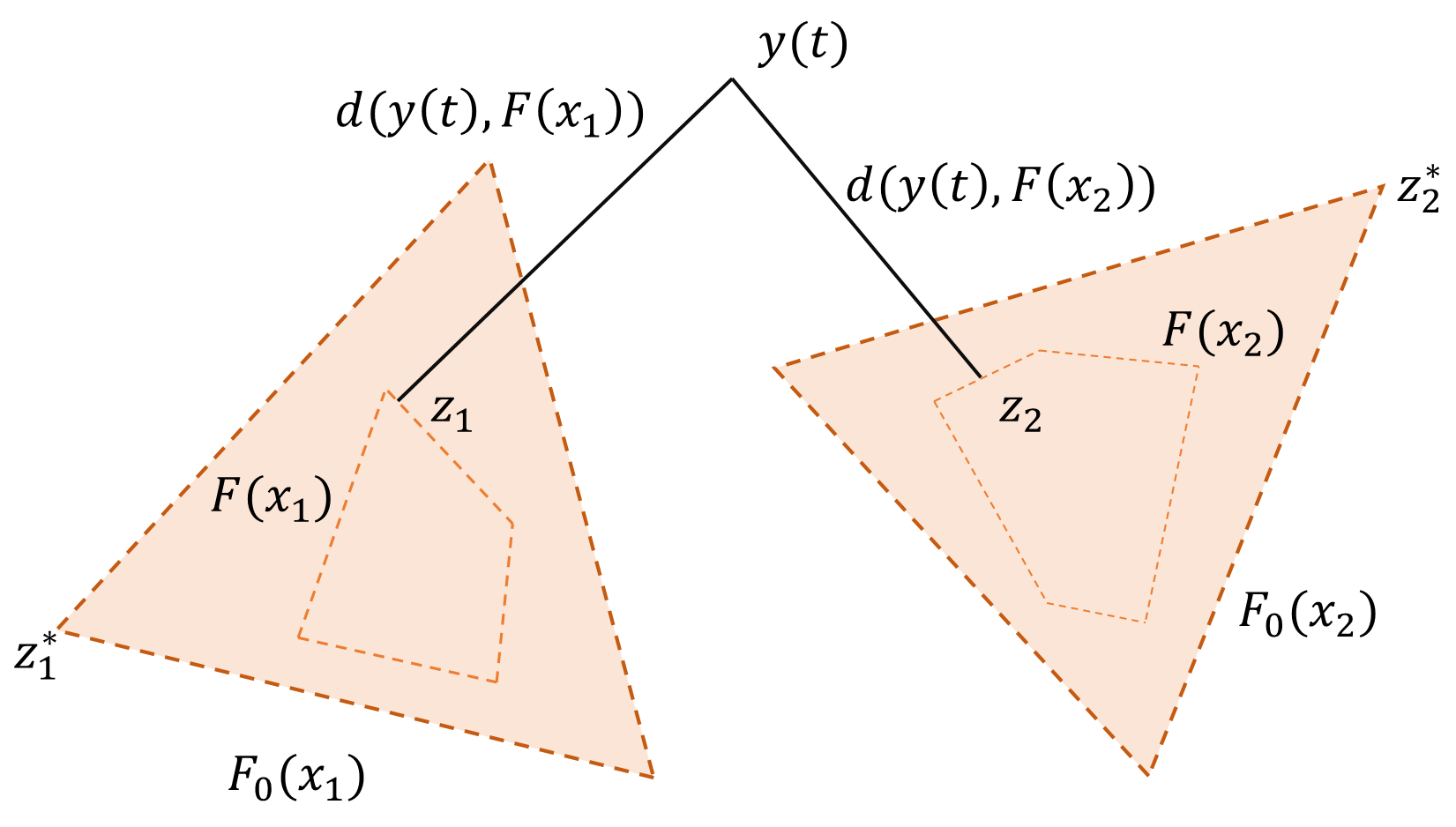}
    \caption{Illustration of the proof that the visibility CBF is Lipschitz continuous in a 2D environment where the pursuer and evader states are positions and orientations. The triangles show the pursuer's unoccluded FoV, while the dashed polygons inside show the occluded FoV. The distances to the points $\bfz_1$ and $\bfz_2$ from $\bfy(t)$ define the visibility CBF at $\bfx_1$ and $\bfx_2$, respectively.}
    \label{fig:Lipschitz_proof}
\end{figure}
 
\begin{proof}
Consider two points $(t_1, \bfx_1)$ and $(t_2, \bfx_2)$. Then, by the triangle inequality:
\begin{align*}
|h(t_1, \bfx_1) \!&-\! h(t_2, \bfx_2)| = |d(\bfy(t_1), \calF(\bfx_1)) \!-\! d(\bfy(t_2), \calF(\bfx_2))|\\
& \leq |d(\bfy(t_1), \calF(\bfx_1)) - d(\bfy(t_1), \calF(\bfx_2))|\\ 
& \qquad + |d(\bfy(t_1), \calF(\bfx_2)) - d(\bfy(t_2), \calF(\bfx_2))|.
\end{align*}
The distance from a point to a fixed set is 1-Lipschitz \cite{distance_to_set}, and by the mean value theorem, it follows that
\begin{align*}
|d(\bfy(t_1), \calF(\bfx_2))  - d(\bfy(t_2), \calF(\bfx_2))|&\leq \|\bfy(t_1) - \bfy(t_2)\|\\ 
&\leq k|t_1 - t_2|.
\end{align*}
Let $\bfz_1 \in \pa \calF(\bfx_1)$ and $\bfz_2 \in \pa \calF(\bfx_2)$ be such that $d(\bfy(t_1), \calF(\bfx_1)) = \|\bfy(t_1) - \bfz_1\|$ and $d(\bfy(t_1), \calF(\bfx_2)) = \|\bfy(t_1) - \bfz_2\|$. Then,
\[
|d(\bfy(t_1), \calF(\bfx_1)) - d(\bfy(t_1), \calF(\bfx_2))| \leq \|\bfz_1 - \bfz_2\|.
\]

With slight abuse of notation, let $\calF_0(\bfx) = A(\bfx)\calF_0+b(\bfx)$ be the unoccluded FoV at pursuer state $\bfx$. Since $\calF(\bfx) \subseteq \calF_0(\bfx)$, we have $\bfz_1 \in \calF_0(\bfx_1)$ and $\bfz_2 \in \calF_0(\bfx_2)$, and thus:
\begin{align*}
\|\bfz_1 - \bfz_2\| &\leq m_H(\calF_0(\bfx_1), \calF_0(\bfx_2)) := \!\!\!\!\!\!\sup_{\bfa \in \calF_0(\bfx_1), \bfb \in \calF_0(\bfx_2)} \!\!\!\!\!\! \|\bfa-\bfb\|.
\end{align*}

Now, consider $\bfz_1^* \in \calF_0(\bfx_1)$ and $\bfz_2^* \in \calF_0(\bfx_2)$ such that $\|\bfz_1^*-\bfz_2^*\| = m_H(\calF_0(\bfx_1), \calF_0(\bfx_2))$. Since $\bfz_1^*, \bfz_2^*$ are points obtained from a Lipschitz linear transformation from $\bfx_1, \bfx_2$, there exists a constant $C$ such that
\[
\|\bfz_1^* - \bfz_2^*\| \leq C\|\bfx_1 - \bfx_2\|.
\]
Putting everything together, we conclude that:
\[
|h(t_1, \bfx_1) - h(t_2, \bfx_2)| \leq k|t_1 - t_2| + C\|\bfx_1 - \bfx_2\|. \qedhere
\]
\end{proof}

Fig. \ref{fig:Lipschitz_proof} illustrates the above analysis for a 2D environment with a triangular unoccluded FoV. Thus, based on Def.~\ref{nbf}, we can define a visibility CBF constraint as:
\begin{equation} \label{eq:visibility-nbf}
\inf_{(w,\bfv) \in \partial h(t,x)} \{ \bfv^\top(f(\bfx)+G(\bfx)\bfu) + w\} \geq -\alpha_v(h(t,\bfx)) + \delta,
\end{equation}
where $\alpha_v$ is a Lipschitz extended class-$\calK$ function and $\delta \in \bbR$ is a slack variable that allows relaxing or tightening the visibility constraint. It can also be noted that for points where $h(t,\bfx)$ is differentiable, the above constraint reduces to a usual CBF constraint:
\begin{align} \label{eq:visibility-cbf}
\nabla_{\bfx} d(\bfy&, {\calF(\bfx)})^\top(f(\bfx) + G(\bfx)\bfu) + \nabla_{\bfy} d(\bfy, {\calF(\bfx)})^\top \dot{\bfy}\notag \\ 
&  \leq \alpha_v(-d(\bfy, {\calF(\bfx)})) + \delta.
\end{align}

To capture the safety requirement as well, we use the signed distance function $d(\bfx, \calO)$ from the pursuer state $\bfx$ to the obstacle set $\calO$ to define a CBF. When $d(\bfx,\calO)$ is positive, the pursuer is outside of the obstacle set $\calO$. Similar to the visibility CBF, $d(\bfx, \calO)$ is non-smooth but Lipschitz \cite{distance_to_set}. Thus, we can define a safety CBF constraint as:
\begin{equation}\label{eq:safety-nbf}
\inf_{\bfv \in \partial d(\bfx,\calO)}\{ \bfv^\top (f(\bfx)+G(\bfx)\bfu)\} \geq -\alpha_s(d(\bfx,\calO)),
\end{equation}
which simplifies for differentiable points of $d(\bfx,\calO)$ to:
\begin{equation}\label{eq:safety-cbf}
\nabla_{\bfx} d(\bfx, \calO)^\top(f(\bfx) + G(\bfx)\bfu)  \geq - \alpha_s(d(\bfx, \calO)).
\end{equation}

In summary, we can impose the visibility and safety CBF constraints in \eqref{eq:visibility-nbf} and \eqref{eq:safety-nbf} on a reference control signal $r(t,\bfx,\bfy)$ for the pursuer by synthesizing control inputs as:
\begin{equation}\label{eq:safety-visibility-control}
\min_{\bfu \in \calU, \delta \in \bbR} \; \|\bfu - r(t,\bfx,\bfy)\|^2 + \lambda \delta^2 \quad \mathrm{s.t.} \;\; \eqref{eq:visibility-nbf}, \eqref{eq:safety-nbf},
\end{equation}
where $\delta$ is a slack variable penalized by $\lambda > 0$ that ensures feasibility of the program by relaxing the visibility constraint in \eqref{eq:visibility-nbf} if necessary. Note that since $\partial h(t,\bfx)$ and $\partial d(\bfx,\calO)$ are convex sets (Def.~\ref{def:generalized_gradient}), the infima over linear functions in \eqref{eq:visibility-nbf}, \eqref{eq:safety-nbf} are concave functions of $\bfu$ and, hence, \eqref{eq:visibility-nbf}, \eqref{eq:safety-nbf} are convex constraints in $\bfu$. At points where $h(t,\bfx)$ and $d(\bfx,\calO)$ are differentiable, the constraints are linear in $\bfu$.

\subsection{Planning and Control with Visibility \& Safety Constraints}
\label{sec:planning}
The formulation in \eqref{eq:safety-visibility-control} is myopic in the sense that the future motion of the evader is predicted only through its instantaneous velocity $\dot{\bfy}(t)$. To maintain visibility of the evader, it is desirable to synthesize a non-myopic control policy for the pursuer that takes into account a longer-horizon prediction of the evader motion.

One option is to formulate an optimal control problem with horizon $T$ to obtain a non-myopic control policy $\pi(t,\bfx,\bfy)$:
\begin{equation*}
\min_{\pi, \delta} \int_t^{t+T} \|\pi(\tau,\bfx(\tau),\bfy(\tau)) - r(\tau,\bfx(\tau),\bfy(\tau))\|^2 + \lambda \delta^2(\tau) d\tau
\end{equation*}

subject to the visibility and safety constraints in \eqref{eq:visibility-nbf}, \eqref{eq:safety-nbf} and the pursuer dynamics in \eqref{eq:pursuer}. Unfortunately, there are a number of challenges with solving this problem. First, it requires predicting the evader configuration $\bfy(\tau)$ over a long horizon $\tau \in [t,t+T)$, which is much more challenging than estimating the instantaneous evader velocity $\dot{\bfy}(t)$ as required in \eqref{eq:safety-visibility-control}. Second, the constraints in \eqref{eq:visibility-nbf} and \eqref{eq:safety-nbf} would no longer be convex in $\bfu$ because, at times $\tau > t$, the policy $\pi(\tau, \bfx(\tau),\bfy(\tau))$ depends on the pursuer state $\bfx(\tau)$, which evolves nonlinearly according to the dynamics in \eqref{eq:pursuer}. Third, while it is possible to formulate the problem as a nonlinear program, e.g., using CasADi \cite{casadi}, the non-convex and potentially non-smooth constraints \eqref{eq:visibility-nbf}, \eqref{eq:safety-nbf} would make solving it numerically challenging and inefficient.

To avoid the optimal control formulation above, we introduce a trajectory planner that provides a reference control signal $r(\tau)$ for $\tau \in [t,t+T)$ that can be tracked by the evader using the myopic but convex optimization problem in \eqref{eq:safety-visibility-control}. The planner's objective is to rapidly find a collision-free and feasible control trajectory for the pursuer that achieves visibility of the evader as a goal condition.  We use Stable Sparse RRT (SST) \cite{sst}, which is a kinodynamic planner that samples within the control space $\calU$ to generate a reference $r(\tau)$ for $\tau \in [t,t+T)$. To ensure that the visibility constraint is satisfied, the goal condition of SST is set such that $\bfy(t) \in \calF(\bfx(T))$. Note that this goal condition requires the evader configuration $\bfy(t)$ only at time $t$ and, thus, avoids predicting the future evader motion but can still lead to non-myopic behavior via re-planning.

To increase the efficiency of SST, we use the pre-trained Motion Planning Transformer (MPT) \cite{mpt} to suggest sampling regions. MPT uses a transformer architecture that maps patches of an occupancy grid map to a sequence of regions that may contain feasible trajectories to the goal. SST's sampling is biased to the sequence of regions predicted by MPT to focus the tree expansion towards areas that may observe the evader.

A challenge with the formulation in \eqref{eq:safety-visibility-control} is the computation of gradients for the visibility and safety CBFs. In the case of camera or LiDAR sensors, the visibility CBF value can computed by approximating the FoV $\calF(\bfx)$ through ray tracing and finding the distance from the evader configuration $\bfy$ to the FoV. However, there is no analytical method to compute the FoV gradient. We approximate the gradient of the visibility CBF $h(t,\bfx) = - d(\bfy(t),\calF(\bfx))$ using a least-squares finite-difference method. Let $d^*(\bfx) := d(\bfy, \calF(\bfx))$ be the distance to the FoV as a function of the pursuer's state, and let $\calX_p = \{ \delta \bfx_i\}_{i = 1}^N$ be a set of pursuer state perturbations. By Taylor expansion, the distance perturbation is $d^*(\bfx + \delta\bfx_i) \approx d^*(\bfx ) + \nabla_{\bfx} d^*(\bfx)^\top \delta\bfx_i$. Thus, by defining $\Delta d_i^*(\bfx) = d^*(\bfx + \delta\bfx_i) - d^*(\bfx )$, we can approximate the gradient as $\nabla_{\bfx} d^*(\bfx) \approx \arg\min_{\bfp} \sum_{i=1}^{N} \| \delta\bfx_i^\top \bfp - \Delta d^*_i(\bfx)\|$, which is computed using least squares. To achieve higher gradient accuracy, a larger number of perturbations $N$ should be considered. For points where $d^*(\bfx)$ is not differentiable the least-squares problem does not have a unique solution. Generalized gradient vectors can be obtained as $\bfv = \bfC^\dagger \bfc + (I-\bfC^\dagger \bfC)\bfn$, where $\bfC\in \bbR^{N\times n}$ has rows $\delta\bfx_i^\top$, $\bfc \in \bbR^N$ has elements $\Delta d^*_i(\bfx)$, and $\bfn \in \bbR^n$ is arbitrary.

The safety CBF value and gradient can be estimated from a point cloud observation, $\calP = \{\bfrho_i\}_{i=1}^P$ with $\bfrho_i \in \bbR^p$, e.g., obtained from a LiDAR or RGBD camera sensor. Consider the distance and direction of $N$-nearest points in $\calP$ to the pursuer, determined as $\psi_j = \| \bfrho_j-\phi(\bfx)\|$ and $\bfg_j = (\bfrho_j - \phi(\bfx))/\| \bfrho_j-\phi(\bfx)\|$, respectively, where the function $\phi: \bbR^n \to \bbR^p$ maps the pursuer state $\bfx$ to its position $\phi(\bfx) \in \bbR^p$. The generalized gradient can be approximated as the set of these directions vectors, i.e. $\partial g(\bfx) \approx \{\bfg_j\}_{j=1}^N$, and the safety CBF value can be estimated as the minimum distance, $d(\bfx,\calO) \approx \min_{j\in\{1\cdots N\}}\psi_j$.

\section{EXPERIMENTS AND RESULTS}
\label{sec:experiments}

We use a mobile robot, equipped with a camera and a LiDAR, as a pursuer in our evaluation. The robot has state $\bfx = [x, y, \theta]^\top \in \mathbb{R}^2 \times [-\pi, \pi)$ consisting of its position $(x,y)$ and orientation $\theta$, input $\bfu = [v, \omega]^\top \in \mathbb{R}^2$ defining its linear and angular velocities, and differential-drive dynamics, $\dot{x} = v\cos(\theta)$, $\dot{y} = v\sin(\theta)$, $\dot{\theta} = \omega$. In our experiments, the environment consists of obstacles $\calO \subset \mathbb{R}^2 \times [-\pi, \pi)$ placed on a planar surface. We use Hector SLAM \cite{HectorSLAM} with 2D scans extracted from the LiDAR point clouds for occupancy mapping and pursuer state localization. The evader is modeled as a system with position $\bfy \in \bbR^2$ and velocity $\dot{\bfy} \in \bbR^2$. The pursuer uses camera detections of the evader and an extended Kalman filter \cite{kalmanbucy1961} with constant velocity motion model to estimate the evader state. The FoV $\calF(\bfx)$ is computed using ray tracing in the occupancy map generated by Hector SLAM.  The visibility and safety CBF values and gradients are computed as discussed at the end of Sec.~\ref{sec:planning}. We also include box constraints $\calU$ on $\bfu$ in \eqref{eq:safety-visibility-control}. We evaluated the performance of our method in 3D CARLA simulations \cite{Dosovitskiy17} and on a real Jackal robot.

\subsection{CARLA Simulations}

\begin{figure}
    \centering
    \begin{subfigure}[b]{0.9\linewidth}
        \includegraphics[width=\textwidth]{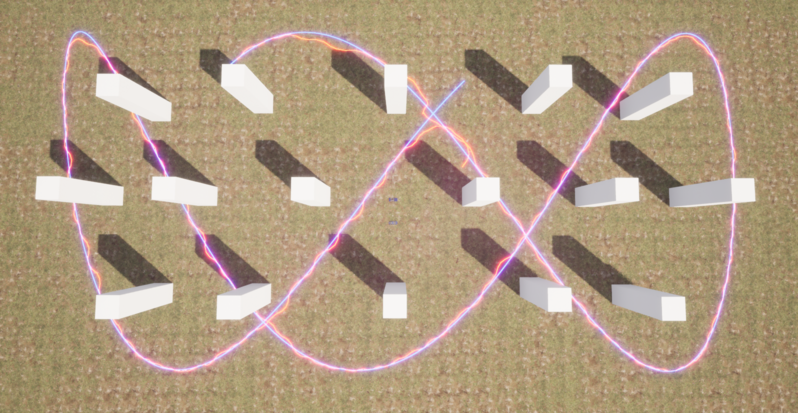}
        \caption{Pursuer (red) and evader (blue) trajectories in the CARLA simulation. The environment spans $400 \times 400$ $m^2$ and has sixteen white pillars of size $5 \times 5$ $m^2$ as obstacles.}
        \label{fig:Carla_env}
    \end{subfigure}
    \begin{subfigure}[b]{\linewidth}
        \includegraphics[width=\textwidth]{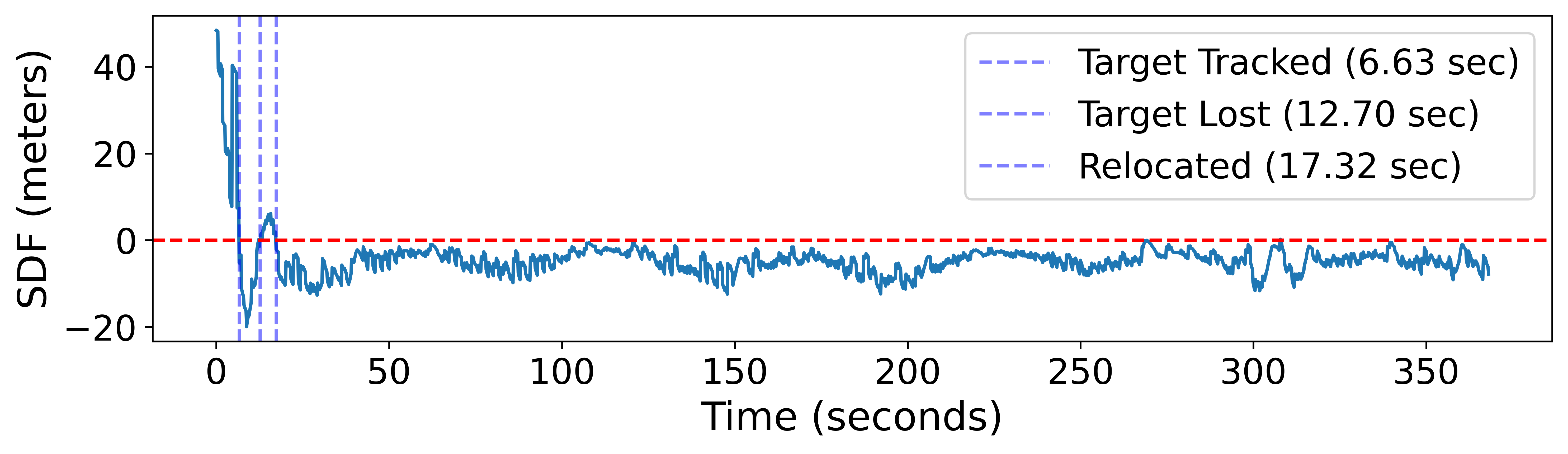}
        \caption{SDF values from the evader to the pursuer's FoV. The red line denotes the FoV boundary. Segments above the line indicate that the evader is outside the FoV. The initial distance between the evader and pursuer was 110 meters.}
        \label{fig:CARLA_everything_stats}
    \end{subfigure}
    \caption{CARLA simulation trajectory}
    \label{fig:Carla_trajectories}
\end{figure}

A 3D environment and two cars that act as pursuer and evader were simulated in the CARLA autonomous driving simulator \cite{Dosovitskiy17}, shown in Fig. \ref{fig:Carla_env}. The pursuer's FoV $\calF_0$ was a circular sector with range $80$ m spanning $60^\circ$. The control space $\calU$ was set to $[0, 12]$ m/s $\times$ $[-1, 1]$ rad/s. The evader follows a Lissajous curve, executing periodic movements within the simulation environment, described by $\bfy(t)=\left[A\sin(at+\gamma), B\sin(bt)\right]^\top$, with $A=180$, $a=0.15$, $B=90$, $b=0.40$, and $\gamma=2.05$. The evader's path includes straight sections and sharp turns close to obstacles, allowing us to assess the controller's capabilities in tracking and obstacle avoidance. Measurements of $\bfy(t)$ and $\dot{\bfy}(t)$ were provided to the pursuer directly by the CARLA simulator.

\begin{figure}
    \centering
    \begin{subfigure}[b]{0.32\linewidth}
        \includegraphics[width=\linewidth]{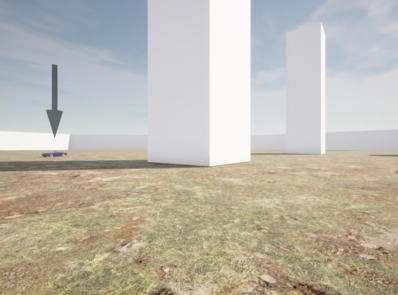}
    \end{subfigure}%
    \hfill%
    \begin{subfigure}[b]{0.32\linewidth}
        \includegraphics[width=\linewidth]{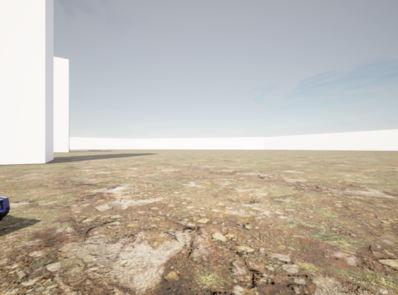}
    \end{subfigure}%
    \hfill%
    \begin{subfigure}[b]{0.32\linewidth}
        \includegraphics[width=\linewidth]{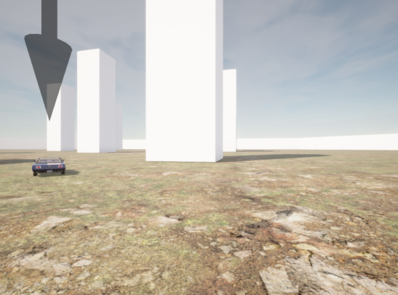}
    \end{subfigure}\\
    \begin{subfigure}[b]{0.32\linewidth}
        \includegraphics[width=\linewidth]{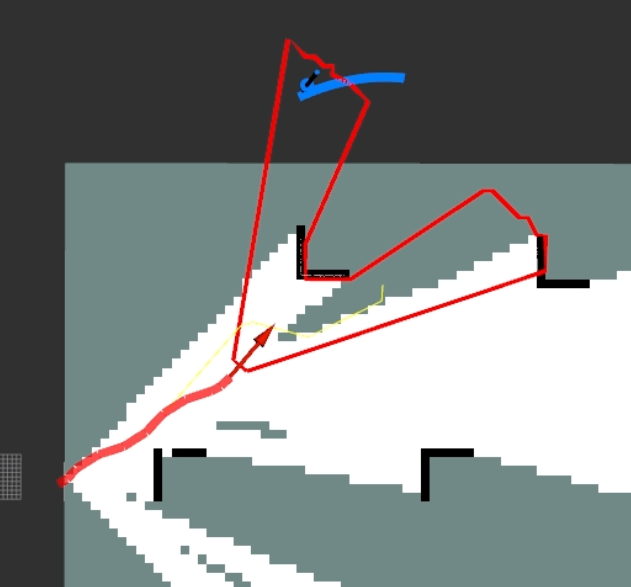}
        \caption{Target tracked}
        \label{fig:Target_Tracked}
    \end{subfigure}%
    \hfill%
    \begin{subfigure}[b]{0.32\linewidth}
        \includegraphics[width=\linewidth]{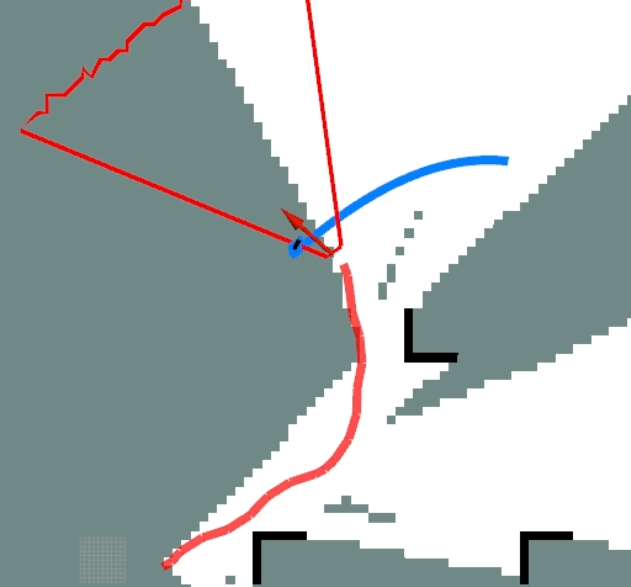}
        \caption{Target lost}
        \label{fig:Target_Lost}
    \end{subfigure}%
    \hfill%
    \begin{subfigure}[b]{0.32\linewidth}
        \includegraphics[width=\linewidth]{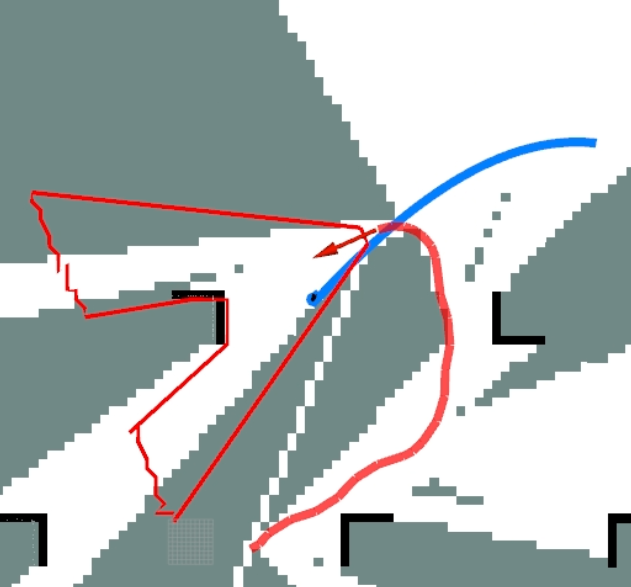}
        \caption{Relocated}
        \label{fig:Relocalization}
    \end{subfigure}
    \caption{Top: Snapshots from pursuer's view along its trajectory. Bottom: The red and blue curves denote the pursuer and evader trajectories, respectively. (a) The instance when the target is first observed by the pursuer; (b) The pursuer loses the evader around a corner; (c) The pursuer relocates the evader after turning the corner.}
    \label{fig:CARLA_snapshots}
\end{figure}

In the simulation setup, shown in Fig. \ref{fig:Carla_env}, the evader starts outside of the pursuer's FoV. Initially, the pursuer lacks evader and environment information. Thus, the pursuer’s trajectory until the first evader detection is not included in the assessment. As shown in Fig. \ref{fig:Target_Tracked}, pursuer achieves its first evader detection within 11 seconds and loses tracking only once due to occlusion, as shown in Fig. \ref{fig:Target_Lost}. It took 4.6 seconds to re-establish tracking, as depicted in Fig. \ref{fig:Relocalization}. The SDF values shown in Fig. \ref{fig:CARLA_everything_stats} indicate that the pursuer successfully tracked the evader with brief loss in visibility during sharp turns or near obstacles.

\begin{table}
\centering
\caption{Performance metrics for simulated ablations in CARLA}\label{table:carla_results} 
\resizebox{\linewidth}{!}{%
\begin{tabular}{cccc}\hline
\textbf{Ablation} & Planner only & Controller only & \textbf{Full system}\\ \hline
\textbf{Initialization time} $\downarrow$ & 24 s & 8 s  & \textbf{7 s}\\ 
\textbf{\% of time in FoV} $\uparrow$ & 59\% & 97\%  & \textbf{98\%}\\
\textbf{Mean SDF} $\downarrow$ & 8.3 m & -3.9 m  & \textbf{-5.0 m}\\ 
\textbf{Max. relocate time} $\downarrow$ & 37 s & 6.4 s & \textbf{4.6 s}\\
\textbf{No. of Collisions} $\downarrow$ & 4 & \textbf{0} & \textbf{0}\\
\textbf{Control Frequency} $\uparrow$ & 5 Hz & \textbf{37 Hz} & 32 Hz\\
\hline
\end{tabular}}
\end{table}

We also carried out an ablation study of the planner and controller components in our method with different pursuer and evader initialization. We compared: (a) directly using the planned reference trajectory to control the pursuer (\emph{Planner only}); (b) using only the controller with visibility and safety CBF constraints (\emph{Controller only}); and (c) Asynchronous planner+controller (\emph{Full system}). The results are reported in Table \ref{table:carla_results}. \emph{Planner only} struggled without real-time safety constraints, leading to collisions. \emph{Controller only} performed better than \emph{Planner only}, maintaining visibility and avoiding collisions. \emph{Full system} shows the best performance in visibility and safety across all metrics, maintaining the evader within the pursuer's FoV for 98\% of the trajectory while avoiding collisions with obstacles. This shows our method’s effectiveness for reliable, safe navigation in realistic simulated environments.

\subsection{Real World Experiments}

In the real experiments, the pursuer was a differential-drive Jackal robot, equipped with an OS-1 Ouster 3D LiDAR and D455 RealSense depth camera. We used a human-controlled Turtlebot, affixed with an AprilTag as an evader. We imposed a limited, triangular FoV $\calF_0$ which spans $30^\circ$ with range $2$ m to demonstrate our method's ability to deal with limited sensing. We ran the \emph{Controller Only} method at 50 Hz onboard the Jackal, consisting of sensing, mapping, evader detection, and control. Due to the complex environment and high computational demands (about ten times slower for MPT inference), we excluded the planner on Jackal robot, which struggles with incomplete and noisy maps to provide a reliable path. The pursuer control space $\calU$ was set to $[0, 0.5]$ m/s $\times$ $[-0.5, 0.5]$ rad/s. We conducted all experiments in a lab environment and used common objects as obstacles.

\begin{figure}
    \vspace{-1ex}
    \centering
    \includegraphics[width=\linewidth]{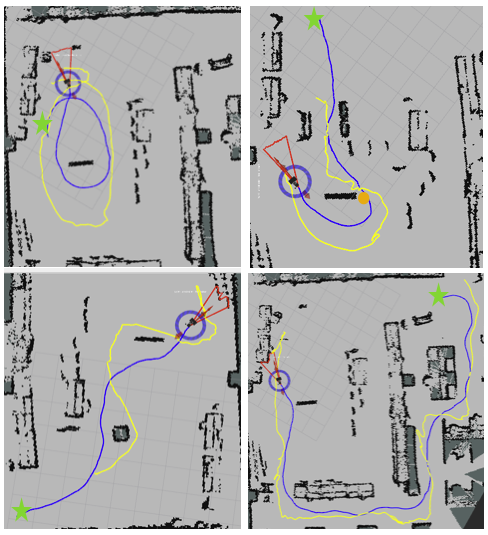}
    \caption{\textbf{Real-world experiments.} The pursuer and evader paths are shown in blue and yellow, respectively. The green star indicates the pursuer's starting position. The red region shows the pursuer FoV. Different experiments are shown clockwise from top left: (1) Simple loop around a single obstacle; (2) Evader follows an 'S' shape. The pursuer grazes an obstacle near a sharp turn denoted by the orange circle; (3) The evader charts a long path through a cluttered environment; (4) The pursuer cuts off the evader.}
    \label{fig:rviz_screens}
\end{figure}

\begin{figure}
    \centering
    \includegraphics[width=\linewidth]{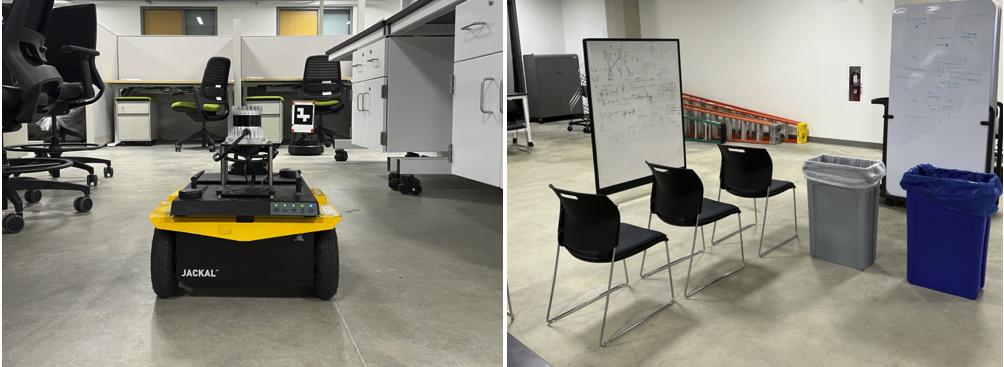}
    \caption{Actual setup for 'S'-shape (left) and a section of long cluttered path (right). On the left, the pursuer navigates through rows of desks. The evader can be seen in the background to the right of the pursuer. On the right, the pursuer navigates through an aisle and turns two corners in an 'S-shape'.}
    \label{fig:combined_real_setup}
    \vspace{1ex}
\end{figure}

\begin{figure}
    \centering
    \includegraphics[width=\linewidth]{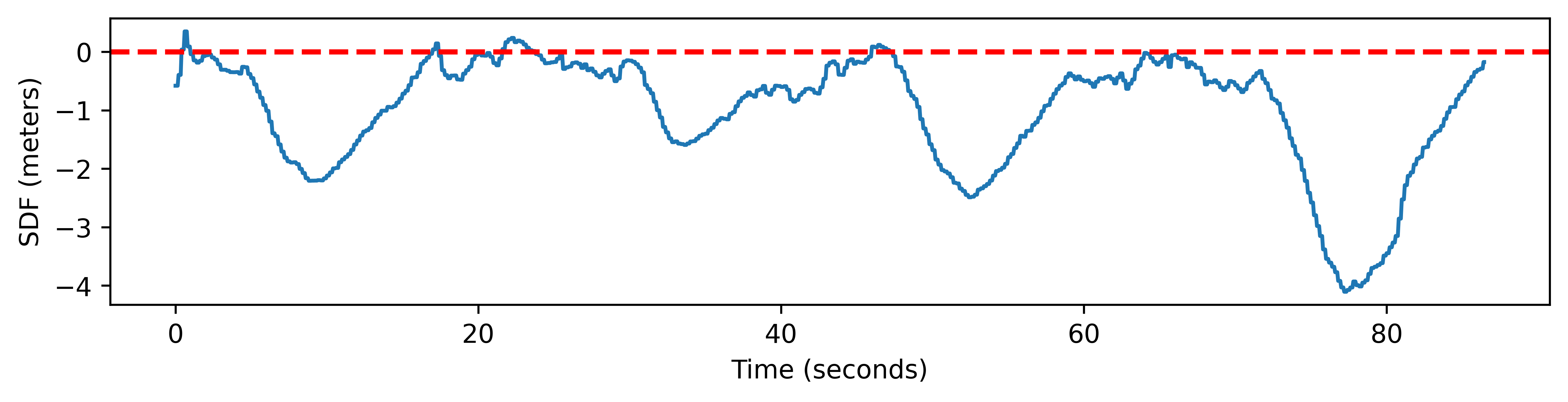}%
    \vspace*{-1ex}
    \caption{SDF of evader to pursuer FoV in Experiment 3 ('Long cluttered path'). The red line corresponds to the FoV boundary. Segments above the line indicate that the evader is outside the FoV.}
    \label{fig:sdf_record}
\end{figure}

\begin{table}
\vspace{2ex}
\centering
\caption{Performance metrics for Jackal robot experiments}\label{results_table} 
\begin{tabular}{ccccc}
\hline \textbf{Experiment} & 1 & 2 & 3 & 4 \\ \hline
\textbf{\% of time in FoV$ \uparrow$} & 92\% & 76\% & 92\% & 91\% \\
\textbf{Mean SDF $\downarrow$} & -0.8 m & -0.25 m & -1.0 m & -1.4 m \\
\textbf{Max. relocate time $\downarrow$} & 1.0 s & 2.9 s & 2.2 s & 0.6 s\\
\textbf{Min. dist. to obstacles $\downarrow$} & 11 cm & 0 cm* & 3 cm & 20 cm\\
\hline
\end{tabular}
\end{table}

We present 4 different experiments, shown in Fig. \ref{fig:rviz_screens}, designed to highlight a specific capability or challenge that our method faces. Fig. \ref{fig:combined_real_setup} shows the actual setup for experiments 2 and 3. We initialized all experiments with the evader inside the pursuer FoV to provide an initial detection for the estimator. Table \ref{results_table} shows performance metrics for each experiment. In experiment 2 (`S-shape'), the pursuer grazes an obstacle (leg of a standing whiteboard) as it turns the corner marked by the orange circle in Fig. \ref{fig:rviz_screens}. The confined space and sharp turns explain lower visibility for this configuration. In experiment 4 (`Cut-off'), we see that the pursuer is able to cut off the evader as it moves around an obstacle, rather than following its path exactly.  Fig. \ref{fig:sdf_record} shows that in experiment 3 ('Long cluttered path'), the SDF value increases during turns as the pursuer momentarily loses the evader, and decreases during straight sections. The max relocate time is the longest time elapsed between losing the evader and bringing it back into the pursuer's FoV. Experiment 4 ('Cut-off') has the lowest relocate time as the pursuer keeps its distance and does not follow the evader around obstacles, a behavior that would cause it to lose sight of the evader for longer. As expected, this metric is the highest for experiment 2 (`S-shape') in which sharp turns mean that the pursuer loses the evader for longer periods. Overall, we see that the pursuer achieves close tracking across various configurations.

\section{CONCLUSION}

This paper introduces a convex programming formulation for control synthesis with visibility and safety constraints. We show that both constraints can be formulated using non-smooth but Lipschitz continuous CBFs. Coupling our controller with a kinodynamic planner that generates non-myopic trajectories to observe a mobile evader, we demonstrate autonomous robot pursuit in 3D simulations and real-world environments using onboard sensing under occlusion. Future work will focus on supplementing our method with an exploration strategy if the pursuer does not begin with a detection of the evader, and with a robust estimator that maintains multiple hypotheses for evader motion.








\clearpage\balance
\bibliographystyle{cls/IEEEtran.bst}
\bibliography{bib/IEEEabrv.bib,bib/ref.bib}

\end{document}